\def\year{2022}\relax
\documentclass[letterpaper]{article} 
\usepackage{aaai22}  
\usepackage{times}  
\usepackage{helvet}  
\usepackage{courier}  
\usepackage[hyphens]{url}  
\usepackage{graphicx} 
\usepackage{natbib}  
\usepackage{caption} 
\DeclareCaptionStyle{ruled}{labelfont=normalfont,labelsep=colon,strut=off} 
\usepackage{algorithm}
\usepackage{algorithmic}
\usepackage{amsmath}
\usepackage{multirow}
\usepackage{makecell}
\usepackage{newfloat}
\usepackage{listings}
\floatstyle{ruled}
\newfloat{listing}{tb}{lst}{}
\floatname{listing}{Listing}
\title{Adversarial Training with Contrastive Learning in NLP}
\author {
    Daniela N. Rim,\textsuperscript{\rm 1}
    DongNyeong Heo, \textsuperscript{\rm 1}
    Heeyoul Choi \textsuperscript{\rm 1} \textsuperscript{\rm 2}
}
\affiliations {
    \textsuperscript{\rm 1} \textit{Handong Global University} - Republic of Korea\\
    \textsuperscript{\rm 2} \texttt{heeyoul@gmail.com}
}

\usepackage{bibentry}

\begin{document}

\maketitle

\begin{abstract}
For years, adversarial training has been extensively studied in natural language processing (NLP) settings. The main goal is to make models robust so that similar inputs derive in semantically similar outcomes, which is not a trivial problem since there is no objective measure of semantic similarity in language. 
Previous works use an external pre-trained NLP model to tackle this challenge, introducing an extra training stage with huge memory consumption during training. 
However, the recent popular approach of contrastive learning in language processing hints a convenient way of obtaining such similarity restrictions. The main advantage of the contrastive learning approach is that it aims for similar data points to be mapped close to each other and further from different ones in the representation space. 
In this work, we propose \textit{adversarial training with contrastive learning} (ATCL) to adversarially train a language processing task using the benefits of contrastive learning. The core idea is to make linear perturbations in the embedding space of the input via fast gradient methods (FGM) and train the model to keep the original and perturbed representations close via contrastive learning. In NLP experiments, we applied ATCL to language modeling and neural machine translation tasks. The results show not only an improvement in the quantitative (perplexity and BLEU) scores when compared to the baselines, but ATCL also achieves good qualitative results in the semantic level for both tasks without using a pre-trained model.
\end{abstract}

\section{Introduction}

Adversarial training has undoubtedly become one of the trending topics in deep learning research in the last decade, since adversarial examples were proposed \cite{szegedy2013intriguing, goodfellow2014explaining}. Adversarial examples can be defined broadly as \textit{samples specifically crafted to fool a neural model but not a human observer}. There are several straightforward ways to craft adversarial examples in computer vision when the model weights and parameters are accessible for the attacker (white-box attack). Some well-known examples based on gradient information are FGSM \cite{goodfellow2014explaining}, BIM \cite{kurakin2016adversarial}, MI-FGSM \cite{dong2018boosting}, JSMA \cite{papernot2016limitations}, and DeepFool  \cite{moosavi2016deepfool}. All these methods result in a corrupted version of an original image, and they are indistinguishable to human observers. After crafting the adversarial examples with one of these methods, the easiest way to make the model robust is to add them to the training set. This adversarial training has a similar effect as regularization by data augmentation \cite{goodfellow2014explaining}.

In contrast to the computer vision applications, adversarial examples are not well defined in natural language processing (NLP) settings. For example, character-level targeted attacks such as character swapping, insertion or deletion, are always perceptible to humans, but are still considered a type of adversarial examples \cite{ebrahimi2018adversarial}. Even though there are many definitions of adversarial examples in the natural language field, some works follow a more general definition, analogous to the computer vision's: \textit{samples specifically crafted to fool the model while preserving the semantic of the original phrase} as in \citeauthor{alzantot2018generating} \citeyear{alzantot2018generating}, \citeauthor{jin2020bert} \citeyear{jin2020bert}, \citeauthor{ren2019generating} \citeyear{ren2019generating}, \citeauthor{morris2020second} \citeyear{morris2020second}, and \citeauthor{garg2020bae} \citeyear{garg2020bae}. This is the definition we consider in this work, and so we only focus on performing imperceptible perturbations in the word-level.

Since the text tokens belong to a discrete space, directly applying small perturbations is unfeasible. As an alternative, we can apply them in the embedding space \cite{cheng2018towards, cheng2019robust}.
The main goal is to make the model robust to small variations of the input so that small variations of the input should not result in different outcomes. 
Prior works introduced a pre-trained NLP model to ensure a strong regularization that makes close representations of the original and perturbed samples \cite{cheng2018towards, cheng2019robust}. However, the errors of the pre-trained model can cause unstable training, and the desired robustness may not be achieved \cite{zhang2020adversarial}. Furthermore, obtaining good adversarial examples via such methods might be computationally expensive, since we have to run the auxiliary model for each sample that we perturbed. 

As an alternative, forcing the representation of similar embeddings (the original and its adversarial perturbation) to be mapped into similar outputs may suggest the usage of contrastive learning. Contrastive learning \cite{chopra2005learning} is a training approach popular in the computer vision field, which aims to bring representations of similar class or instances closer in the representation space, and move them further from different ones. 
With the success of contrastive learning in the computer vision field, there is an increasing interest in applying this method to NLP tasks. However, there are some limitations mainly related (again) to the definition of similarity and dissimilarity in language \cite{rethmeier2020long} in the representation space.

In this work, we propose a simple way of adversarial training for an NLP model with contrastive learning, {\em adversarial training with contrastive learning} (ATCL). In the proposed method, the adversarial examples are generated with a perturbation on embeddings using a Fast Gradient Method (FGM) described in \cite{miyato2016adversarial}. Then, by contrastive learning, the representations of the adversarial and original embeddings are forced to be closer in the higher layer of the network architecture, while some random samples from the minibatch should be represented far from the original one. 


We conducted experiments using the proposed method, ATCL in language model (LM) and neural machine translation (NMT) tasks. ATCL shows general improvements in the perplexity of the language model task, as well as qualitative improvements in the prediction of tokens. Furthermore, ATCL provides better semantic predictions after an adversarially perturbed token compared to the baseline. We also show some qualitative improvement in the NMT task as well, where our model's translation captures better the semantics of the source sentence than the baseline. Our contribution mainly focuses on the use of a contrastive loss, which is simpler to train than an external pre-trained model and shows empirically the preservation of the semantics in the targeted task. 

\section{Preliminaries}

\subsection{Adversarial Examples in NLP} \label{sectionadv}

Given an input sentence $\mathbf{s}$ as a sequence of tokens, $\mathbf{s}=\{x_1, x_2, ... , x_T\}$, we map each discrete token to an embedded representation in the continuous space by $\mathbf{E}x_i = \mathbf{e_i}$. Analogous to \cite{miyato2016adversarial}, we can provide a small linear perturbation in the embedding of a token $\mathbf{e_i}$ to generate an adversarial perturbation $\mathbf{e'_i}$ by:
\begin{equation}\label{miyato}
\mathbf{e'_i} = \mathbf{e_i} -\epsilon \frac{\mathbf{g}}{\|\mathbf{g}\|_2},
\end{equation}
where $\mathbf{g} = \nabla_{\mathbf{e_i}} J(\mathbf{s}, \theta)$ and $J$ is the cost function of $\mathbf{s}$ with parameter $\theta$. If $\mathcal{L}(\mathbf{s}, \theta)$ is the objective function of our original task like cross-entropy, we add a new loss term in the total loss such that:
\begin{equation}
    \mathcal{J}(\theta) = \sum_{\mathbf{s}}\mathcal{L}(\mathbf{s}, \theta) + \alpha  \sum_{\mathbf{s'}}\mathcal{L}_{adv}(\mathbf{s'}, \theta),
\end{equation}
where $\mathcal{L}_{adv}$ is the loss with the adversarial sentence $s'$ composed of adversarial perturbations $\mathbf{e'_i}$, and $\alpha$ is a hyperparameter between 0 and 1.

In the language setting, the total cost function $\mathcal{J}(\theta)$ needs one more restriction. Since the adversarial example results from a small perturbation in the embedding of a token, it will be beneficial if their representations stay close after forwarding them through the model. Without such restriction, the original embedding and its adversarial example might be mapped to completely different outcomes, since the adversarial perturbation is performed in the direction to maximize this effect \cite{cheng2018towards, cheng2019robust}. To address this issue, the resulting objective function for adversarial training can take the following form for extra regularization:
\begin{multline} \label{cost}
        \mathcal{J}(\theta) = \\ \sum_{\mathbf{s}, \mathbf{s'}} \bigg( \mathcal{L}(\mathbf{s}, \theta) +  \alpha  \mathcal{L}_{adv}(\mathbf{s'}, \theta) + \beta \mathcal{L}_{rep}(\mathbf{s}, \mathbf{s'}; \theta) \bigg),
\end{multline} 
 where $\mathcal{L}_{rep}$ is the loss that makes the representations of the original $\mathbf{s}$ and adversarial $\mathbf{s'}$ samples closer after forwarding through the model. The challenge of this objective function is to find a convenient training strategy to compute $\mathcal{L}_{rep}$.

\subsection{Contrastive Learning}

Contrastive learning (CL) has gained popularity in the last few years in the field of computer vision. The main idea is to train the representation layer of a model by a contrastive loss objective function \cite{chopra2005learning} to pull closer representations of the same class field (i.e. original image and its augmented one(s) called positive samples) and separate them from the rest of the images (negative samples). 
In a self-supervised setting \cite{wu2018unsupervised, henaff2020data, hjelm2018learning, tian2020contrastive, chen2020simple, he2020momentum}, given a set of samples $A$, the contrastive loss is calculated by:
\begin{multline} \label{cl}
    \mathcal{L}_{cont} = - \sum_{\mathbf{a}_i \in A} \log \frac{ \exp(\mathbf{a_i} \cdot \mathbf{p_{a_i}} / \tau)}{\sum_{\mathbf{n_a} \in A -\{\mathbf{a}_i\}} \exp(\mathbf{a_i} \cdot \mathbf{n_a} /\tau)},
\end{multline}
where $\mathbf{a_i}$ is the original input or anchor point, $\mathbf{p_{a_i}}$ and $\mathbf{n_a}$  are positive and negative samples, respectively, $(\text{  } \cdot \text{  })$ is the inner product of two vectors, and $\tau$ is a temperature coefficient. The anchor, positives and negatives are processed suitably for the task. For example, in \cite{khosla2020supervised} they are normalized projected representations. 

The main issues of applying CL to the language field for the same purpose as in image settings are the difficulty of establishing a general strategy for text `augmentation', as well as determining many negative samples of an anchor point \cite{rethmeier2021primer}. Despite having less popularity than the image counterpart \cite{jaiswal2021survey} CL in NLP has been used to pre-train zero-shot predictions \cite{rethmeier2020long, pappas2019gile}, and also to improve specific tasks such as language modeling \cite{logeswaran2018efficient, giorgi2020declutr}, text summarization \cite{duan2019contrastive}, and many others \cite{rethmeier2021primer}. 
In the previous approaches, the core purpose of the contrastive learning is to obtain more accurate representations or embeddings of texts/sentences for task-specific purposes.

\section{Proposed approach: Adversarial Training with Contrastive Learning} \label{proposed}

We propose a simple idea to train a language model (or any other language related model like neural machine translation model) with adversarial examples in the embedding space on-the-fly by \textit{FGM}, with the introduction of contrastive loss to ensure that the representations of original tokens and adversarial perturbations are kept close. We call this approach Adversarial Training with Contrastive Learning (ATCL).

First, to generate adversarial examples in language, we put adversarial perturbation on the representation of words, especially complete words excluding symbols, numbers and subwords. Let $\mathbf{S} = \{\mathbf{s}_1, \mathbf{s}_2, ... , \mathbf{s}_B\}$ be a batch of sentences, where $\mathbf{s}_k = \{x_{k1}, x_{k2}, ... , x_{kN}\}$ is an input sentence to the model consisting of $N$ tokens $x_{ki}$, and $\mathbf{E s}_k = \{\mathbf{E}x_{k1}, \mathbf{E}x_{k2}, ... , \mathbf{E}x_{kN} \} = \{\mathbf{e}_{k1}, \mathbf{e}_{k2}, ... , \mathbf{e}_{kN} \}$ the sentence embedding which is a sequence of word embeddings. To exclude non-complete words for adversarial examples, we define a mask $\mathcal{M}$ by $\mathcal{M}(\mathbf{\mathbf{Es}_k}) = \{\mathcal{M}(\mathbf{E}x_{ki})\}$, where:
\[ 
\mathcal{M}(\mathbf{E}x_{ki})=\begin{cases} 
      1 & \text{if } x_{ki}\in \mathcal{V}_R, \\
      0 & \text{otherwise},
      \end{cases}
\]
where $\mathcal{V}_R$ is the restricted subset of tokens from the vocabulary $\mathcal{V}$. Specifically, $\mathcal{V}_R$ will not contain tokens of lonely characters (e.g. \textit{`a'}, \textit{`Z.'}), symbols (e.g. `!', `@', `\#'), and numbers. If the vocabulary comes from Byte-Pair Encoding \cite{sennrich2015neural}, $\mathcal{V}_R$ would also leave out segmented words or subwords. After applying this mask $\mathcal{M}$ to the embedded sentence $\mathbf{Es}_k$, we randomly choose a candidate $\mathbf{e}_{kj}$ from $\mathbf{s}_k$, such that $\mathcal{M}(\mathbf{E}x_{kj}) \neq 0$. The purpose of $\mathcal{M}$ is to prevent the model from choosing some meaningless tokens as candidates for adversarial perturbation. Finally, we obtain an adversarial sample $\mathbf{e}'_{kj}$ from $\mathbf{e}_{kj}$ by Eq. \ref{miyato}.

\begin{figure}[ht]
    \centering
    \includegraphics[width=0.474\textwidth]{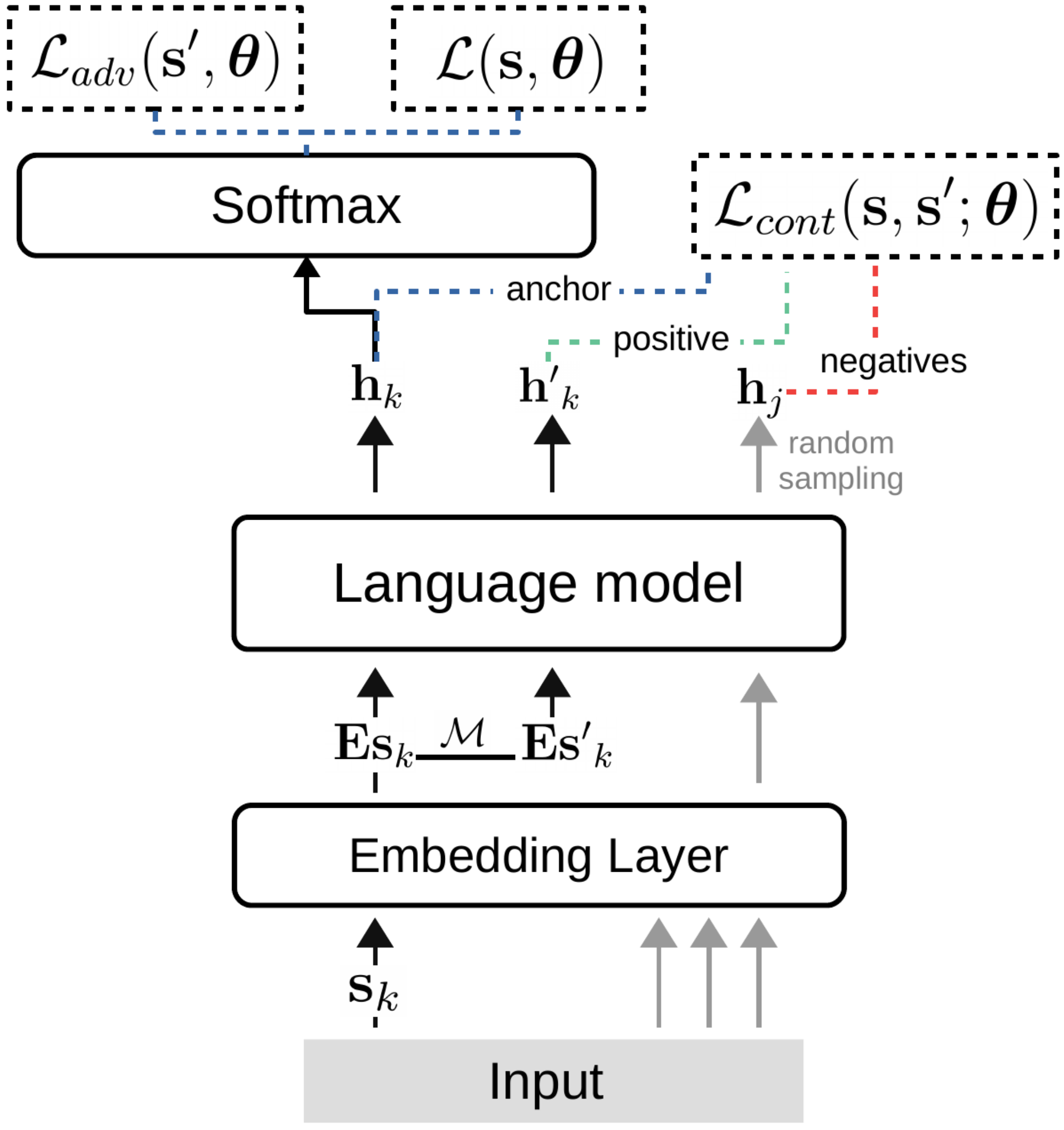}
    \caption{Basic architecture for our proposed approach. A batch $\mathbf{S}$ of sentences $\mathbf{s}_k$ are given as an input. After the embedding layer we compute the perturbed example in one of the randomly selected tokens from the mask $\mathcal{M}$. Both sentences are forwarded through the main model (in this case, language model.) Right before softmax, we apply contrastive loss to the representations of the forwarded sentences (original one $\mathbf{h}$ and perturbed one $\mathbf{h'}$), and randomly sampled negatives from other sentences in the batch. Finally, we compute the three loses in Eq. \ref{ofcont} accordingly.}
	\label{fig:fig1}
\end{figure}

As shown in Figure \ref{fig:fig1}, the original sentences and their adversarial counterparts (i.e., sentences in which the embedding of a token has been perturbed) are forwarded 
through the encoder to obtain the representation $\mathbf{h}_{k}$ and $\mathbf{h}'_{k}$ right before the softmax layer. These representations have enough information to predict the next word, and they should have the same outcome, since they only differ in a small variation of one token embedding \cite{hahn-choi-2019-self}. However, the adversarial perturbation will move the representation away from the one of original sentence, we need to force the proximity of the representation using contrastive learning. 

Taking the representation of the original candidate $\mathbf{h}_{kj}$ as the anchor point and its adversarial perturbed example $\mathbf{h}'_{kj}$ as the positive, we define the contrastive loss like Eq. \ref{cl}:
\begin{equation} 
    \mathcal{L}_{cont} = \sum_{\mathbf{h_{kj}} \in \mathbf{H(S)}} \mathcal{L}_{kj},
\end{equation}
    where
\begin{equation}\label{lcont}
    \mathcal{L}_{kj} = - \log \frac{ \exp(cosim(\mathbf{h_{kj}}, \mathbf{h'_{kj}})/\tau)}{\sum_{\mathbf{h}_n \in \mathbf{NB}} \exp(cosim(\mathbf{h_{kj}}, \mathbf{h}_{n} )/\tau)},
\end{equation}
where $cosim(\cdot ,\cdot )$ is the cosine similarity of the representations, $\mathbf{NB_{kj}} \equiv \mathbf{H(S)}-\{h_{kj}\}$ are the batches of the sentence representations without the anchor point token's, and $n$ is the number of negatives randomly sampled from this batch $\mathbf{H(S)}$. By sampling negatives from $\mathbf{H(S)}-\{\mathbf{h_{kj}}\}$, we make sure that the anchor point is not selected as a negative sample during each computation. 

Finally, the objective of the ATCL is then computed by 
\begin{equation} \label{ofcont}
    \mathcal{J}(\boldsymbol{\theta})_{ATCL} = \sum_{\mathbf{s}, \mathbf{s'}} \bigg( \mathcal{L} + \alpha  \mathcal{L}_{adv} + \beta \mathcal{L}_{cont} \bigg),
\end{equation} 
which is basically the same as Eq. \ref{cost}, except that the adversarial examples are generated differently and the loss term $\mathcal{L}_{cont}$ is given by Eq. \ref{lcont}, considering language issues as described above. 


\section{Related work}
To the best of our knowledge, adversarial training with contrastive learning has not been successful for natural language in general \cite{jaiswal2021survey}. 

Simultaneous to the development of our work, \cite{chen2021disentangled} proposed using the benefits of contrastive loss for adversarial training. In their work, they showed a disentangled contrastive learning method without using negative samples, and the experiments were carried out only with the BERT architecture. The main difference with our approach is the application of the contrastive learning method: the adversarial perturbations come from backtranslation models instead of an FGM, and the role of the negatives samples is replaced by the joint optimization of two sub-tasks (feature alignment and feature uniformity). Moreover, as the authors mentioned, their approach has some limitations that it works only with their contrastive sub-tasks that needs to be further explored in future works. Although some idea of their work may be shared with our approach, the execution of the method is completely different. Furthermore, our method provides a more general strategy for contrastive learning in adversarial settings that can be applied to any network architectures, not only BERT related tasks.

For ATCL, the challenge lies in the way of computing $\mathcal{L}_{rep}(\mathbf{h}, \mathbf{h'})$ for contrastive learning. The closest approach to our implementation is by \cite{cheng2018towards}, where they used a CNN model as a discriminator \cite{kim2014convolutional} to compute the loss $\mathcal{L}_{rep}$. Another example is \cite{cheng2019robust}, which uses doubly adversarial examples, that is, the restriction in the representations is made by simultaneously training the encoding and decoding of the adversarial example using pre-trained language models. In this case, the loss $\mathcal{L}_{rep}$ is divided into two terms, for the source and the target languages, respectively. One of the drawbacks of these approaches is the additional training of these models, and the fact that they have to be pre-trained by other neural network architecture apart from the original task network. Our method computes this loss without any external pre-trained architecture and take advantage of the benefits of contrastive loss in a rather \textit{naive} way. Our results show that this approach is beneficial to training models for language tasks.

\section{Experiment}

We experimented the proposed idea for language modeling (LM) and neural machine translation (NMT) tasks. The experiments were conducted using at core the Transformer model \cite{vaswani2017attention}. For both tasks we computed only one adversarial sentence $\mathbf{s'}$ per sentence $\mathbf{s}$ in the corpus. With adversarial sentence we refer to a sentence in which one of the embedded tokens has been replaced with a perturbed one.

\subsection{Language Model}
Language models (LMs) based on neural networks capture the syntactic and semantic language regularities \cite{Bengio2003nnlm,Mikolov2010}.
Thus, LMs can predict the probability of tokens $x_i$ given the predecessor sequence $\{x_j\}_{1<j<i-1}$ as conditional probability $p(x_i|x_1,...,x_{i-1})$, or can calculate the likelihood of a given sentence. The likelihood which is a joint probability can be calculated with conditional probability as follows:
\begin{eqnarray}
p(\mathbf{s}) &=& p(x_1, x_2, ... , x_T) \\
&=& p(x_1)\prod_{i=2}^{T} p(x_i|x_1,...,x_{i-1}). 
\end{eqnarray}

We used the architecture of the Tensorized Transformer proposed by \cite{ma2019tensorized} which replaces the standard multi-head attention layers in the original Transformer with a Multi-linear attention. 
The Tensorized Transformer has significantly fewer parameters, in addition to achieving a better Perplexity (PPL) score than the original Transformer \cite{vaswani2017attention}.
We experimented in the Penn TreeBank \cite{mikolov2011empirical} and WikiText-103 \cite{merity2016pointer} datasets as summarized in Table \ref{tab:lmdata}.

\begin{table}[h]
\begin{center}
\begin{tabular}{c|ccc}
    & \textbf{Train}      & \textbf{Valid}      & \textbf{Test}      \\ \hline
    \textbf{PTB}         & 929k       & 73k        & 82k    \\
    \textbf{Wiki-103} & $\sim$103M & $\sim$218k & $\sim$246k \\
\end{tabular}
\centering
\caption{Number of tokens for training/validation/test in the Penn TreeBank (PTB) and WikiText-103 (Wiki-103).}
\label{tab:lmdata}
\end{center}
\end{table}

For both datastets, the hyperparameters $\epsilon$ (Eq. \ref{miyato}), $\alpha$, $\beta$ (Eq. \ref{cost}) and the temperature coefficient $\tau$ (Eq. \ref{lcont}) were set to 0.03, 0.1, 0.1, and 0.07, respectively, and the batch size was 45.

\subsubsection{Quantitative Results}

For quantitative comparison, the perplexity scores of the LM task are presented in Table \ref{tab:lmresults}. To evaluate the effects of adversarial training and contrastive learning separately, on the Tensorized Transformer (TT) `Baseline TT', we added adversarial training (`+Adv. only' which means $\beta = 0$ in Eq. \ref{ofcont}) and ATCL (`+ATCL'). 
In general, `+ATCL' improves the PPL scores significantly compare to the others. Note that `+Adv. only' does not improve for Penn Tree Bank, though it does a little for Wikitext 103, which means that contrastive learning works effectively on the adversarial examples.  

To check the effect of the number of negative samples in the contrastive loss (i.e., $n$ in Eq. \ref{lcont}.), we experimented with different amount of randomly sampled negatives from the batches. 
We can see that increasing the number is not always beneficial. For the WikiText-103 dataset, increasing the number of randomly sampled negatives gives a better PPL score. However, for a smaller dataset like PTB the adding more negatives harms the performance. We believe that the effect of the number of negative samples depends on the amount of unique tokens in the batch. PTB dataset has less number of unique tokes, which makes selecting ``bad" negatives very plausible (tokens with no significant semantic impact).

\begin{table*}[t]
\begin{tabular}{l|cccc}
\hline
\multirow{2}{*}{\textbf{Model and Training}} & \multicolumn{2}{c}{\textbf{Wikitext 103}} & \multicolumn{2}{c}{\textbf{Penn Tree Bank}} \\ \cline{2-5} 
 & \textbf{Validation}         & \textbf{Test}        & \textbf{Validation}          & \textbf{Test}         \\ \hline
\textbf{T with adaptive input}*   &  19.80  &  20.50  &  59.10  & 57.00 \\ 
\textbf{T XL base}* & 23.10 & 24.00 & 56.72  & 54.52 \\ 
\textbf{T XL +TT} * & 23.61 &  25.70 & 57.90  & 55.40  \\ \hline
\textbf{Baseline TT} & 22.70 & 22.42 & 41.91 & 36.13 \\ 
\textbf{Baseline +Adv. only} & 21.75 & 21.67 & 42.68 & 36.46 \\ \hline
\textbf{Baseline +ATCL} (n=5) & 22.79 & 22.59 & 37.93 & 32.89\\ 
\textbf{Baseline +ATCL} (n=10) & 21.75 & 21.67 & \textbf{35.29} & \textbf{29.08} \\  
\textbf{Baseline +ATCL} (n=20) & \textbf{20.73} & \textbf{20.61} & 42.52 & 36.85 \\ 
\hline
\end{tabular}
	\centering
	\caption{Perplexity (PPL) of language model for the validation and test datasets. \textbf{T} stands for Transformer \cite{vaswani2017attention} and \textbf{TT} for Tensorized Transformer \cite{ma2019tensorized}. We experimented based on our implementation, comparing them with other results using Transformer architectures given by \cite{ma2019tensorized} (*). We experimented our adversarial setting with (+ATCL) and without contrastive loss (+Adv. only), and in parenthesis we report the amount of negatives randomly sampled from the batch.}
	\label{tab:lmresults}
\end{table*}

\subsubsection{Word Representation}

One of the advantages of contrastive learning is that it brings closer similar samples in the representation space, while pushing them further apart from different ones. We study this effect qualitatively in the text corpus.

After training, we computed the closest neighbors (by Euclidean distance) to the anchor words. In Table \ref{tab:lmneigh} we show the neighbors for three tokens after training the baseline, training only using FGM (i.e., `+Adv. only'), and with our proposed idea (`+ATCL'). Surprisingly, we can observe that contrastive loss helps generate more semantic representation in the word-level. 
For example, for a word with negative connotations such as \textit{\textbf{`hate'}}, its neighbors are negative words after contrastive learning. In contrast, the baseline model and training with only adversarial loss show a mix of synonyms and antonyms as neighbors of the anchor word. This suggests that the contrastive loss has the effect of rearranging the representation space with semantic information. By contrastive loss, words with similar embeddings are forced to be mapped to the same outcome, and at the same time, different representations are forced to be apart. This result in a semantic clustering 
that helps with the word-level semantics.

\begin{table*}[t]
\begin{center}
    \begin{tabular}{c|c|c|c}
    \hline
    \textbf{Word}   & \textbf{Baseline}      & \textbf{+Adv. only  }    & \textbf{+ATCL}    \\ \hline
    \textbf{friend} & \makecell{`brother', `daughter', \\ `understanding', \textbf{`director'}} & \makecell{`cousin', `colleague', \\ \textbf{`knowledge'}, `mentor'} & \makecell{`cousin', `fellow', \\ `partner', `colleague'} \\ \hline
    \textbf{hate}   & \makecell{`admit', `prejudice', \\ `troubled', `regret'} & \makecell{\textbf{`loving'}, `hurt', \\ \textbf{`embrace'}, `committing' } & \makecell{`dirty', `poison', \\ `regret', `shame'} \\ \hline
    \textbf{like }  & \makecell{\textbf{`under'}, `so', \\ \textbf{`offering'}, \textbf{`@.@'}} & \makecell{\textbf{`under'}, `inspired',\\ \textbf{`less'}, \textbf{`led'} } & \makecell{`before', `just', \\ `alongside', `despite' }  \\ \hline
    \end{tabular}
	\caption{Four closest neighbors of some words in the vocabulary of the WikiText-103 dataset after training with the baseline (refer to Table \ref{tab:lmresults}), with only the adversarial loss, and with our proposed training loss (Eq. \ref{lcont}). In \textbf{bold}, we marked the words with different connotation from the anchor word. We observe that the contrastive loss promotes some qualitative improvement in the word-level representation. }
    \label{tab:lmneigh}
\end{center}
\end{table*}

\subsubsection{Robustness Against Attacks}

We also conducted a qualitative study of targeted adversarial attacks. Given a sentence from the test set, we perturbed the embedding of a target word following the method described in Section \ref{sectionadv}. This word is around the end of the sentence and they should be meaningful (we used the mask criteria $\mathcal{M}$ explained before.) 
The set of the perturbed sentences was forwarded through the baseline model and our proposed model, and the models complete the sentence after the perturbed word with the predicted probability for the next words. 

Some generated results are presented in Table \ref{tab:lmgen}, where in most cases, our proposed approach yields better predictions than the baseline model. For example, the first result shows that our model generates nicely ``a starring support role", while the baseline generates ``a starring joint television", which is broken due to the targeted attack. The average perplexities of the generated sentences from the baseline and from our `+ATCL', are 10.12 and 8.77, respectively. Thus, after applying the perturbations, the generated sentences from the ATCL have produced a lower loss in the trained model than the ones generated from the baseline.
Note that even though we trained our model to be resistant of perturbations in sentence-level, the results show that it is also robust against targeted perturbation in the word-level. 
\begin{table*}[t]
\begin{center}
 \begin{tabular}{|l|l|}
\hline
Baseline & \begin{tabular}[c]{@{}l@{}}He had a guest @-@ starring role on the television series The Bill in 2000 . This was followed by \\ a \textbf{starring} \textit{join television.} \end{tabular} \\ \hline
+ATCL    & \begin{tabular}[c]{@{}l@{}}He had a guest @-@ starring role on the television series The Bill in 2000 . This was followed by \\ a \textbf{starring} \textit{support role.} \end{tabular} \\ \hline \hline
Baseline & \begin{tabular}[c]{@{}l@{}}The massive North Korean attack had made deep penetrations everywhere in the division sector \\ except in the north in the zone \textbf{of} \textit{city}. \end{tabular} \\ \hline
+ATCL     & \begin{tabular}[c]{@{}l@{}}The massive North Korean attack had made deep penetrations everywhere in the  division sector \\ except in the north in the zone \textbf{of} \textit{town}. \end{tabular} \\ \hline \hline
Baseline & \begin{tabular}[c]{@{}l@{}}Hall was involved for one month during preproduction; his ideas for lighting the film began with his \\ first reading of the script, and further passes allowed him to refine his approach before \textbf{other} \textit{sunk}. \end{tabular} \\ \hline
+ATCL     & \begin{tabular}[c]{@{}l@{}}Hall was involved for one month during preproduction; his ideas for lighting the film began with his \\ first reading of the script, and further passes allowed him to refine his approach before \textbf{other} \textit{shot}. \end{tabular} \\ \hline \hline
Baseline & \begin{tabular}[c]{@{}l@{}}However , they are rarely considered for a modern design , the principles behind them having been \\ superseded by other methodologies which are more \textbf{accurate} \textit{design that} $<$unk$>$.\end{tabular}  \\ \hline
+ATCL     & \begin{tabular}[c]{@{}l@{}}However , they are rarely considered for a modern design , the principles behind them having been \\ superseded by other methodologies which are more \textbf{accurate} \textit{patent of elements}.\end{tabular} \\ \hline
\end{tabular}
	\caption{Examples of next word prediction after adversarially perturbing the embedding of a token in a sequence. In \textbf{bold}, we marked the words that were perturbed by following the Eq. \ref{miyato}, and in \textit{italic}, the predicted tokens by the differently trained models (baseline and ours.) Our approach completes sentences which make more sense given the context than the baseline.}
    \label{tab:lmgen}   
\end{center}
\end{table*}

\subsection{Neural Machine Translation}

Neural machine translation (NMT) consists of processing tokens in a certain source language and predicting the corresponding sequence of tokens in a target language based on encoder and decoder with end-to-end training \cite{Sutskever2014,Bahdanau2015}. That is, given a sequence $\mathbf{s^{src}} = \{x_1, x_2, ... , x_D\}$, it is mapped to a target sequence $\mathbf{s^{trg}} = \{y_1, y_2, ... , y_M\}$. 
In general, the objective of NMT is to maximize the conditional probability as follows
\begin{equation}
p(\mathbf{s^{trg}}|\mathbf{s^{src}}) = \prod^{M}_{m=1} p(y_m|y_{<m}, \mathbf{s^{src}}; \boldsymbol{\theta}),
\end{equation}
where the parameter $\boldsymbol{\theta}$ is updated to maximize the conditional probability. 

We experimented with the IWSLT'17 datasets for English-German (En-De) and English-French (En-Fr) pairs. Both datasets are segmented using the Byte-Pair Encoding (BPE) method \cite{sennrich2015neural}.  Since we only choose one token to adversarially perturb per sentence, we only considered as adversarial candidates complete word tokens (i.e., not subwords) using the mask $\mathcal{M}$. The details of each language pair dataset are shown in Table \ref{tab:nmtdata}.

\begin{table}[ht]
\begin{center}
\begin{tabular}{c|ccccc}
    & \textbf{Train} & \textbf{Valid}  & \textbf{Test}   & \textbf{Vocab}      & \textbf{BPE tok} \\ \hline
    \textbf{En-De}  & $\sim$160k & $\sim$7k & $\sim$6k & 10k (s) & 36.63\% \\ 
    \textbf{En-Fr} & $\sim$200k & $\sim$7k & $\sim$6k & 28k (ns) & 24.34\% \\
    \end{tabular}
    	\centering
\caption{Details of amount of parallel corpus sentences for En-De and En-Fr training, validation, and test sets. We also show the vocabulary length (and in parenthesis if it is \textbf{s}hared or \textbf{n}ot \textbf{s}hared) and the percentage of subwords in the vocabulary.}
\label{tab:nmtdata}
\end{center}
\end{table}

For the NMT experiments we used the same configurations of \cite{vaswani2017attention} except for the number of layers in the encoder and decoder. Since there was no significant difference in performance, for these datasets we used 3 layers instead of 6.

\subsubsection{Results}
The experiments for the NMT task were carried out in a similar modality as the LM task. We show the result of different configurations for each of the language pairs in Table \ref{tab:nmtresults}. Again, we tried with different number of negatives. 
We can see that ATCL increases the BLEU score compared to the baseline, though the improvement is not as significant as in LM. There is some cases like `Fr-En', training with only the adversarial loss gives a slightly better performance than ATCL. 

\begin{table}[ht]
\begin{center}

\begin{tabular}{|l|c|c|c|c|}
\hline
\textbf{Model name}   & \textbf{En-De} & \textbf{De-En} & \textbf{En-Fr} & \textbf{Fr-En} \\ \hline
\begin{tabular}[l]{@{}l@{}}\textbf{Baseline} \\ \textbf{(Transformer S)}\end{tabular} & 24.61 & 30.34 & 35.23 & 34.51 \\ \hline
\textbf{Baseline +Adv} & 25.04 & 30.36  & 35.02 & \textbf{35.97} \\ \hline
\multirow{3}{*}{\textbf{Baseline +ATCL}}  & 24.63 & \textbf{31.34} & \textbf{36.40} & 35.35 \\ \cline{2-5} 
 & \textbf{25.26} & 30.36 & 35.60  & 35.48 \\ \cline{2-5} 
    & 24.74 & 30.13  & 35.38 & 35.46 \\ \hline
\end{tabular}
	\caption{BLEU test scores for the IWSLT dataset (we used IWSLT15' as testing datasets.). `Baseline + ATCL' has 3 results with 5, 10 and 20 negative samples as in LM.}
    \label{tab:nmtresults}
\end{center}
\end{table}

As shown in Table \ref{tab:nmtresults}, the advantages of using the contrastive loss for NMT task are less visible. We speculate that the main reason of this lack of improvement is the use of BPE. Our approach relies heavily on embedding similarity, i.e. we perform the experiments under the assumption that similarity (closeness) in the word representation space involves semantic similarity. In this case, the training is harmed by this assumption. The representation of subwords due to BPE in the embedding space is less straightforward to understand than in the LM task. Even though we masked the input sentences so that we could only choose complete words, when sampling the negatives from the batch we do not rely on this mask. 
Thus, contrastive loss might be partly based on some meaningless subwords for negative samples which can diminish the advantage of contrastive learning for semantic representation.

We also show some qualitative results in Table \ref{tab:nmttrans}. Given the source (\textbf{src}) and target (\textbf{trg}) sentences, we show some translations for the En-De language pairs using the baseline (\textbf{Baseline}) and our model (\textbf{ATCL}). The latter shows some semantic improvement in the quality of the translations. 

\begin{table*}[ht]
\begin{center}
\begin{tabular}{|c|l|}
\hline
\multicolumn{2}{|c|}{Example from De-En}\\
\hline
\textbf{src (De)} & Und ich dachte mir : ``wow , das muss \textbf{unbedingt} in die filmkunst einbezogen werden." \\ 
\hline
\textbf{trg (En)} & And I thought , ``wow , this is something that needs to be embraced into the cinematic art." \\ \hline
\textbf{Baseline}  & And I thought to myself , ``wow , this may have been reluctant to the movie." \\ 
\hline
\textbf{Adv.} & And I thought , ``wow , this has to be very much involved in filmmaking." \\ 
\hline 
\textbf{ATCL} & and i thought , ``wow , this is something that needs to be involved in film art." \\ 
\hline \hline
\multicolumn{2}{|c|}{Example from Fr-En}\\
\hline
\multirow{2}{*}{\textbf{src (Fr)}} & Voici Toxoplasma gondii, ou Toxo pour les amis, parce \textbf{qu’}une créature terrifiante à toujours droit  à un \\ 
& joli surnom.  \\
\hline
\multirow{2}{*}{\textbf{trg (En)}} & This is Toxoplasma gondii, or Toxo, for short, because the terrifying creature always deserves a cute \\
& nickname. \\ 
\hline
\textbf{Baseline}  &  This is Toxo, or Toxo from friends, because a terrifying creature to a Beautiful nickname.\\ 
\hline
\textbf{Adv.} & This is Toxo, or Toxo from friends, because a terrifying creature is always right for a beautiful nickname. \\ 
\hline 
\textbf{ATCL} & This is Toxo, or Toxo from friends, because a terrifying creature is always right for a beautiful nickname. \\ 
\hline
\end{tabular}
	\caption{Examples of De-En and Fr-En translations after perturbing the embedding of some tokens (in \textbf{bold}) in the source (\textbf{src}). We show the translations of the target sentence (\textbf{trg}), the baseline model (\textbf{Baseline}), the model trained with only adversarial perturbations (\textbf{Adv.}), and our approach (\textbf{ATCL}). These examples show that our model seems to capture better the underlying semantic meaning of the source sentence than the baseline. However, there are some cases in which the results are similar between \textbf{Adv.} and \textbf{ATCL}, also reflected in the BLEU scores results of Table \ref{tab:nmtresults}.}
    \label{tab:nmttrans}
\end{center}
\end{table*}

\section{Conclusion and Future Work}
In this work we proposed a simple idea for combining the advantages of the contrastive learning to adversarial training in language tasks. Our approach was applied to language modeling and neural machine translation tasks with promising results. We show some benefits of forcing the proximity of the representations of the original tokens and their adversarial ones via contrastive learning. We believe this to be a simple approach to train NLP models adversarially without the use of a pre-trained models or any limit on the model architecture.

This implementation can be improved further in some ways. First, random sampling from a batch may not be the optimal strategy for obtaining negatives. Even though we show that for the LM task training with these negatives gives a lower PPL score, we believe that further improvements can be achieved with negatives carefully crafted according to semantics. Another question is if contrastive learning can be applied to representation spaces in which the ``similarity" concept is not easy to interpret, like in language settings, where cosine similarity may be too simple. Future works should include a thorough study on better similarity measures.

Despite these limitations, we believe this to be a first approach to introduce the benefits of contrastive learning into adversarial training for language tasks without the use of external pre-trained models. 




\bibliography{aaai22.bib}

\end{document}